\documentclass{article}
\pdfoutput=1

\usepackage{uai_proceed2e}
\usepackage{times}

\RequirePackage[style=authoryear,maxnames=2, maxbibnames=99, backend=bibtex,
 natbib=true, url=true, eprint=true, doi=false, isbn=false, dashed=false, uniquelist=false]{biblatex}
\bibliography{uai2016.bib}

\usepackage{booktabs}
\usepackage{graphicx}

\usepackage{amsmath}
\usepackage{amssymb}
\usepackage{siunitx}
\usepackage{algorithm}
\usepackage{algpseudocode}

\usepackage[all=normal,lists]{savetrees}

\newcommand\ftnote[1]{\footnote{\raggedright#1}}

\graphicspath{{./figures/}}

\mathchardef\mhyphen="2D
\DeclareMathOperator*{\argmin}{arg\,min}
\DeclareMathOperator*{\argmax}{arg\,max}

\title{Bayesian Hyperparameter Optimization for Ensemble Learning}

\author{ \hspace*{-0.3em}{\bf Julien-Charles L\'{e}vesque\thanks{~~julien-charles.levesque.1@ulaval.ca},~ Christian Gagn\'{e}}\\
Laboratoire de vision et syst\`{e}mes num\'{e}riques\\
D\'{e}p. de g\'{e}nie \'{e}lectrique et de g\'{e}nie informatique\\
Universit\'{e} Laval, Qu\'{e}bec, Canada
\And
{\bf Robert Sabourin}\\
Laboratoire d'imagerie, de vision et d'IA\\
D\'{e}p. de g\'{e}nie de la production automatis\'{e}e\\
\'{E}cole de technologie sup\'{e}rieure, Montr\'{e}al, Canada
}

\begin{document}

\maketitle

\begin{abstract}
In this paper, we bridge the gap between hyperparameter optimization and ensemble learning by performing Bayesian optimization of an ensemble with regards to its hyperparameters. Our method consists in building a fixed-size ensemble, optimizing the configuration of one classifier of the ensemble at each iteration of the hyperparameter optimization algorithm, taking into consideration the interaction with the other models when evaluating potential performances. We also consider the case where the ensemble is to be reconstructed at the end of the hyperparameter optimization phase, through a greedy selection over the pool of models generated during the optimization. We study the performance of our proposed method on three different hyperparameter spaces, showing that our approach is better than both the best single model and a greedy ensemble construction over the models produced by a standard Bayesian optimization.
\end{abstract}

\section{INTRODUCTION}

For a long time, the tuning of hyperparameters for learning algorithms was solved by simple exhaustive methods such as grid search guided by cross-validation error. Grid search does work in practice, but it suffers from serious drawbacks such as a search space complexity that grows exponentially with the number of hyperparameters tuned. 
Recently, other strategies such as sequential model-based parameter optimization~\citep{Hutter2011}, random search~\citep{Bergstra2012}, and Bayesian optimization~\citep{Snoek2012} have been shown to be better alternatives to grid search for non-trivial search spaces.

While hyperparameter optimization focuses on the performance of a single model, it is generally accepted that ensembles can perform better than single classifiers, one of many striking examples being the winning entry of the Netflix challenge~\citep{Bell2007}. More recent machine learning competitions such as Kaggle competitions are also often won by ensemble methods~\citep{Sun2011}. Given these previous results, it is logical to combine Bayesian hyperparameter optimization techniques with ensemble methods to further push generalization accuracy. \citet{Feurer2015nips} performed post-hoc ensemble generation by reusing the product of a completed hyperparameter optimization, winning phase 1 of the ChaLearn AutoML challenge~\citep{Guyon2015}. Lastly, \citet{Snoek2015} also constructed post-hoc ensembles of neural networks for image captioning.

These two lines of previous work make for a compelling argument to directly apply Bayesian optimization of hyperparameters for ensemble learning. Rather than trying to model the whole space of ensembles, which is likely hard and inefficient to optimize, we pose a performance model of the ensemble at hand when adding a new classifier with some given hyperparameters. This is achieved by reusing models previously assessed during the optimization, evaluating performance change induced by adding them one at a time to the ensemble. This allows us to compute observations of the true ensemble loss with regards to the hyperparameter values. These observations are used to condition a Bayesian optimization prior, creating mean and variance estimates over the hyperparameter space which will be used to optimize the configuration of a new classifier to add to the ensemble. Finally, we consider different possibilities to maintain and build the ensemble as the optimization progresses, and settle on a round-robin optimization of the classifiers in the ensemble. This ensemble optimization procedure comes at a very small additional cost compared with a regular Bayesian optimization of hyperparameter yet yields better generalization accuracy for the same number of trained models.

We evaluate our proposed approach on a benchmark of medium datasets for two different hyperparameter spaces, one consisting solely of SVM algorithms with different kernel types, and one larger space with various families of learning algorithms. In both search spaces, our approach is shown to outperform regular Bayesian optimization as well as post-hoc ensemble generation from pools of classifiers obtained by classical Bayesian optimization of hyperparameters. We also evaluate our approach on a search space of convolutional neural networks trained on the CIFAR-10 dataset. The proposed approach is also able to provide better performance in this case.

The paper is structured as follows: Section~\ref{sec:hpo} presents the problem of Bayesian hyperparameter optimization and highlights some related work. Section~\ref{sec:eo} presents the main contributions of this paper, which can be summarized as a methodology for Bayesian optimization of ensembles through hyperparameter tuning. Finally, Section~\ref{sec:exp} presents the experiments and an analysis of the results.

\section{HYPERPARAMETER OPTIMIZATION}
\label{sec:hpo}

The behavior of a learning algorithm $A$ is often tunable with regards to a set of external parameters, called hyperparameters $\gamma = \{\gamma_1, \gamma_2, \dots \} \in \Gamma$, which are not learned during training. The hyperparameter selection problem is one stage of a bi-level optimization problem, where the first objective is the tuning of the model's parameters $\theta$ and the second objective is the performance with regards to the hyperparameters $\gamma$.

The procedure requires two datasets, one for training and one for hyperparameter optimization (also called validation), namely $\mathcal{X}_T$ and $\mathcal{X}_V$, each assumed to be sampled \emph{i.i.d.} from an underlying distribution $\mathcal{D}$. The objective function to minimize for hyperparameter optimization takes the form of the empirical generalization error on $\mathcal{X}_V$:
\begin{gather}
\label{eq:hpo}
    f(\gamma) = L(h_\gamma \vert \mathcal{X}_V) + \epsilon \\
    L(h_\gamma \vert \mathcal{X}_V) = \frac{1}{\vert \mathcal{X}_V \vert} \sum_{i=1}^{\vert\mathcal{X}_V\vert} l_{0-1}(h_\gamma(x_i), y_i),
\end{gather}
where $\epsilon$ is some noise on the observation of the generalization error, $l_{0-1}$ is the zero-one loss function, and the model $h_\gamma$ is obtained by running the training algorithm with hyperparameters $\gamma$, $h_\gamma = A(\mathcal{X}_T, \gamma)$. Other loss functions could be applied, but unless otherwise specified, the loss function will be the zero-one loss.

In order to solve this problem, Bayesian optimization consists in posing a probabilistic regression model of the generalization error of trained models with respect to their hyperparameters ${\gamma}$, and exploiting this model to select new hyperparameters to explore. At each iteration, a model of $f(\gamma)$ is conditioned on the set of previously observed hyperparameter values and associated losses $\{ \gamma_i, L(h_{\gamma_i} \vert \mathcal{X}_V)\}_{i=1}^t$.
Selection of the next hyperparameters to evaluate is performed by maximizing an \emph{acquisition function} $a(\gamma \vert f(\gamma))$, a criterion balancing exploration and exploitation given mean and variance estimates obtained from the model of $f(\gamma)$. Among the model families for $f(\gamma)$, two interesting choices are Gaussian Processes~\citep{Rasmussen2006,Snoek2012} and Random Forests~\citep{Hutter2011}, both providing information about the mean and variance of the fitted distribution over the whole search space.

A typical hyperparameter optimization is executed iteratively, subsequently generating a model of $f(\gamma)$ from observations, selecting hyperparameter tuples $\gamma$ to evaluate, training a classifier $h_\gamma$ with the given training data, evaluating it on the validation data, and looping until the maximum number of iterations or time budget is spent.

Recent advances in hyperparameter optimization have primarily focused on making optimization faster, more accurate and applicable to a wider set of applications. In order to speed up convergence, \citet{Feurer2015aaai} have shown that hyperparameter optimization can be warm started with meta features about datasets. Touching on both speed and optimality, the design of better acquisition functions has seen a lot of interest, and predictive entropy search was shown to be less greedy than expected improvement in locating the optima of objective functions~\citep{Hernandez2014}. New applications have also emerged, one notable example being the optimization of hyperparameters for anytime algorithms with freeze-and-thaw Bayesian optimization~\citep{Swersky2014}.

\subsection{HYPERPARAMETER OPTIMIZATION AND ENSEMBLES}
\label{sec:post_hoc}

The idea of generating ensembles with hyperparameter optimization has already received some attention. \citet{Bergstra2013} applied hyperparameter optimization in a multi-stage approach akin to boosting in order to generate better representations of images. \citet{Lacoste2014smbeo} proposed the Sequential Model-based Ensemble Optimization (SMBEO) method to optimize ensembles based on bootstrapping the validation datasets to simulate multiple independent hyperparameter optimization processes and combined the results with the agnostic Bayesian combination method.

The process of hyperparameter optimization generates many trained models, and is usually concluded by selecting a model according to the hold-out (or cross-validation) generalization error $\gamma^* = \argmin_\gamma L(h_\gamma \vert \mathcal{X}_V)$. This single model selection at the end of the optimization is the equivalent of a point estimate, and it can result in overfitting. One strategy to limit this overfitting in the selection of a final model is to select multiple models instead of one, reducing the risk of overfitting and thus increasing the generalization performance.

A simple strategy to build an ensemble from a hyperparameter optimization is to keep the trained models as they are generated for evaluation instead of discarding them~\citep{Feurer2015nips}. This effectively generates a \emph{pool} of classifiers to combine at the end of the optimization, a process which is called \textbf{post-hoc ensemble generation}. Forward greedy selection has been shown to perform well in the context of pruning a pool of classifiers~\citep{Caruana2004}. At each iteration, given a pool of trained classifiers $H$ to select from, a new classifier is added to the ensemble, selected according to the minimum ensemble generalization error. At the first iteration, the classifier added is simply the single best classifier. At step $t$, given the ensemble $E = \{h_{e_1}, h_{e_2}, \dots, h_{e_{t-1}}\}$, the next classifier is chosen to minimize the empirical error on the validation dataset when added to $E$:
\begin{gather}
    h_{t} = \argmin_{h \in H} L(E \cup \{h\} \vert \mathcal{X}_V) \\
    \label{eq:loss_set}
    L(E \cup \{h\} \vert \mathcal{X}_V) = \sum_{i=0}^{\vert \mathcal{X}_V \vert} l_{0-1}\left(\,\strut g(x_i, E \cup \{h\}), y_i\,\right),
\end{gather}
where $g(x_i, E)$ is a function combining the predictions of the classifiers in $E$ on sample $x_i$. In this case, the combination rule is majority voting, as it is less prone to overfitting~\citep{Caruana2004,Feurer2015nips}. Other possible combination rules include weighted voting, stacking~\citep{Kuncheva2004} and agnostic Bayesian combination~\citep{Lacoste2014agnostic}, to name only a few.  Such an approach can be shown to perform better than the single best classifier produced by the hyperparameter optimization, due in part to a reduction of the classifiers' variance through combination.


\section{ENSEMBLE OPTIMIZATION}
\label{sec:eo}

In this work, we aim at directly optimizing an ensemble of classifiers through Bayesian hyperparameter optimization. The strategies discussed in the previous section mostly aimed at reusing the product of a completed hyperparameter optimization after the fact. The goal is to make an online selection of hyperparameters that could be more interesting for an ensemble, but which do not necessarily maximize the objective function of Equation~\ref{eq:hpo} on their own.
Directly posing a model on the space of all possible ensembles of a given size $f(E) = f(\gamma_1, \dots, \gamma_m)$ would result in a very hard and inefficient optimization problem, effectively duplicating the training of many models.

In order to palliate this, we propose a more focused approach. We define the objective function to be the performance of a given ensemble $E$ when it is augmented with a new classifier trained with hyperparameters $\gamma$, or $h_\gamma$. In other words, the objective function is the empirical error provided by adding a model $h_\gamma$ to the ensemble $E$:
\begin{equation}
\label{eq:ens_opt}
    f(\gamma \vert E) = L (E \cup A(\gamma, \mathcal{X}_T) \vert \mathcal{X}_V),
\end{equation}
again using the empirical loss on a hold-out validation set $\mathcal{X}_V$. Contrarily to the post-hoc ensemble generation, a probabilistic model is fit on the performance that a model trained with given hyperparameters would provide to the ensemble. In order to do this, two things are required: 1) an already established ensemble, and 2) a pool of trained classifiers available to compute \emph{observations} of Equation~\ref{eq:ens_opt} to condition our model on. Given the history of trained models so far $H = \{h_1, \dots, h_t\}$ and an ensemble defined by a selection of classifiers within the history $E = \{h_{e_1}, \dots, h_{e_m}\}$, the observations used to model $f(\gamma)$ are obtained by reusing the trained classifiers in $H$, keeping $E$ constant. Consequently, the objective function models the true ensemble error.
Given a zero-one loss function and an empty ensemble $E = \varnothing$, Equation~\ref{eq:ens_opt} falls back to a classical hyperparameter optimization problem, and the objective function will be minimized by the best hyperparameters for a single model $\gamma^*$.

\begin{figure}[tb]
    \centering
    \includegraphics[width=0.48\textwidth]{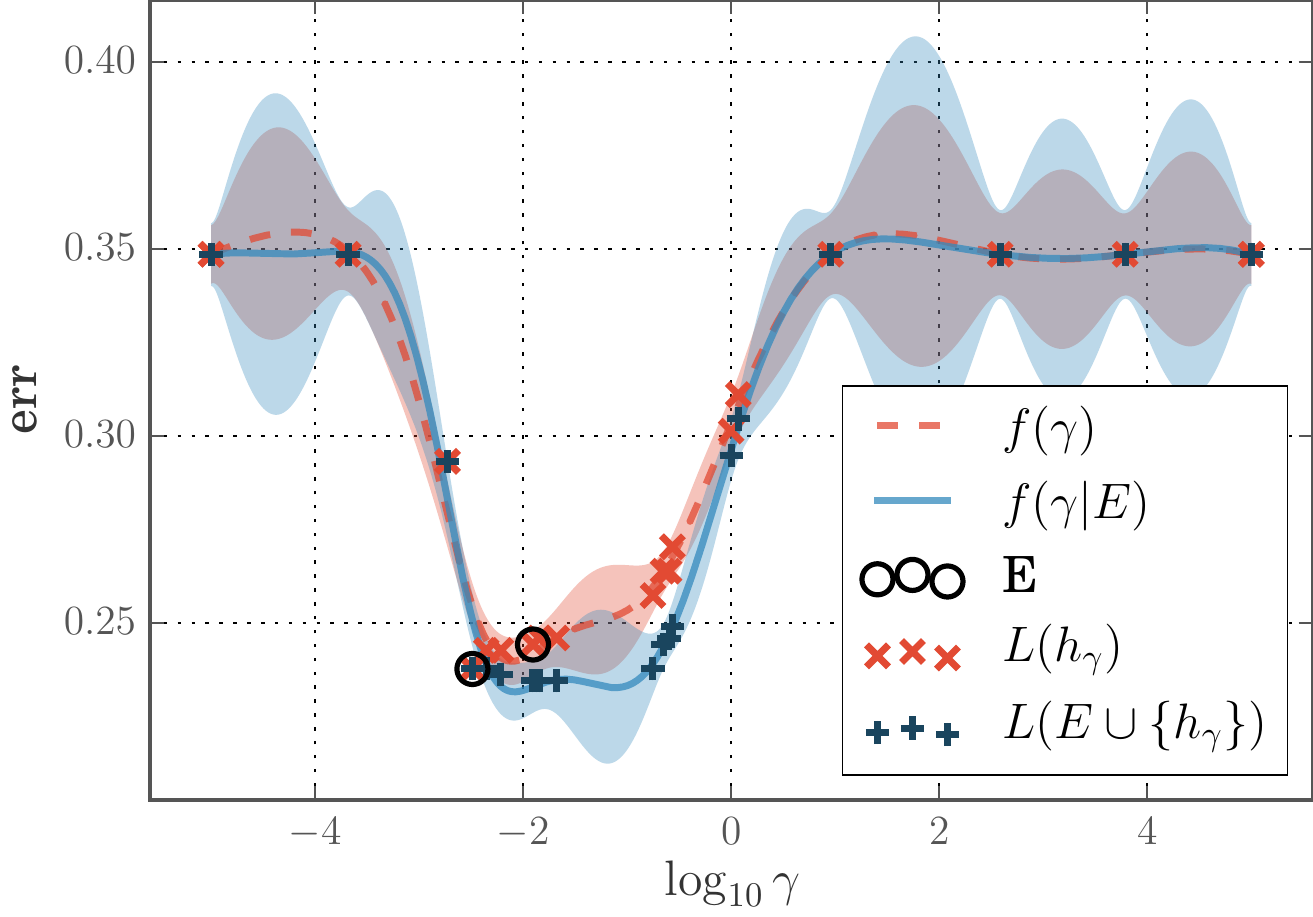}
    \caption{Example of an ensemble optimization. The red marks represent trained models and their standalone generalization error, and black circles represent two models selected for the current ensemble $E$. Blue marks represent the performance of an ensemble when we add the trained model with corresponding hyperparameters $\gamma$.}
    \label{fig:ex_eo}
\end{figure}

The power of the suggested framework is illustrated with an example shown in Figure~\ref{fig:ex_eo}. This figure presents one iteration of ensemble optimization given 20 trained SVMs in a one-dimensional hyperparameter space, where the single hyperparameter is the width of the RBF kernel ($\sigma \in [10^{-5}, 10^5]$). The dataset used is the Pima Indian Diabetes dataset available from UCI~\citep{Frank2010}, with separate training and validation splits. The current ensemble $E$ consists of two models selected by forward greedy selection shown by black circles. Ensemble evaluation and member selection strategies will be discussed further; for now let us assume a fixed ensemble. The red Xs represent the generalization error of single models and the red curve represents a Gaussian Process prior conditioned on those observations, in other words, a model of $f(\gamma)$. The blue crosses represent the generalization error of the ensemble $E$ when the corresponding classifiers are added to it, and the blue curve is again a Gaussian Process prior conditioned on the ensemble observations, or, more generally speaking, a model of $f(\gamma \vert E)$. For both Gaussian Processes, the variance estimates are represented by shaded areas. The next step would be to apply an acquisition function with the ensemble mean and variance estimates to select the next hyperparameters to evaluate.

Figure~\ref{fig:ex_eo} shows that the objective function of an ensemble and a single classifier can be different. It can also be observed in this case that the generalization error of the ensemble is lower than that of a single model, hence the interest in optimizing ensembles directly.

\subsection{ALTERNATE FORMULATIONS}

In order to be able to generalize over the space of hyperparameters, it is crucial to have an ensemble which does not contain all the classifiers in $H$, because if it did there would be no information added in the computation of Equation~\ref{eq:ens_opt}. A different problem formulation could be derived which compares classifiers with the whole pool of trained models, which would take the form $f(\gamma \vert H) = q(h_\gamma \vert H, \mathcal{X}_V)$,
where $q(\cdot)$ is a metric of performance for a classifier with regards to the pool. For example, a diversity inducing metric such as pairwise disagreement~\citep{Kuncheva2004} could be used, but this would lead to degenerate pools of classifiers, as diversity is easily increased by trivial and degenerate classifiers (voting all for one class or the other).

Multi-objective optimization approaches have been considered for the maximization of both diversity and accuracy, a problem typically solved with genetic algorithms~\citep{Tsymbal2005}. However, this problem formulation does not guarantee a better performing ensemble -- only a more diverse pool of classifiers -- with the hope that it will lead to better generalization performance. Directly optimizing diversity in classifier ensembles is questionable, and the evidence thus far is mixed~\citep{Didaci2013,Kuncheva2003elusive}. 

Lastly, an inverse problem could be posed, measuring the difference in the generalization error by removing classifiers from the history one by one, and optimizing this difference. One problem with such a model is that it would be vulnerable to redundancy -- very good hyperparameters present in multiple copies in the history would be falsely marked as having no impact on the generalization error.

For the reasons stated above, the best solution appears to be the use of a fixed ensemble which is maintained and updated as the optimization progresses. Thus it is possible to build an accurate Bayesian model of how well an ensemble would perform if we added a model trained with hyperparameters $\gamma$. This means that we need to store the trained classifiers in a database (or store their predictions on the validation and testing splits) to permit ensemble construction and evaluation in the subsequent iterations.

\subsection{ENSEMBLE UPDATE}

\begin{figure}[tb]
    \centering
    \includegraphics[width=0.48\textwidth]{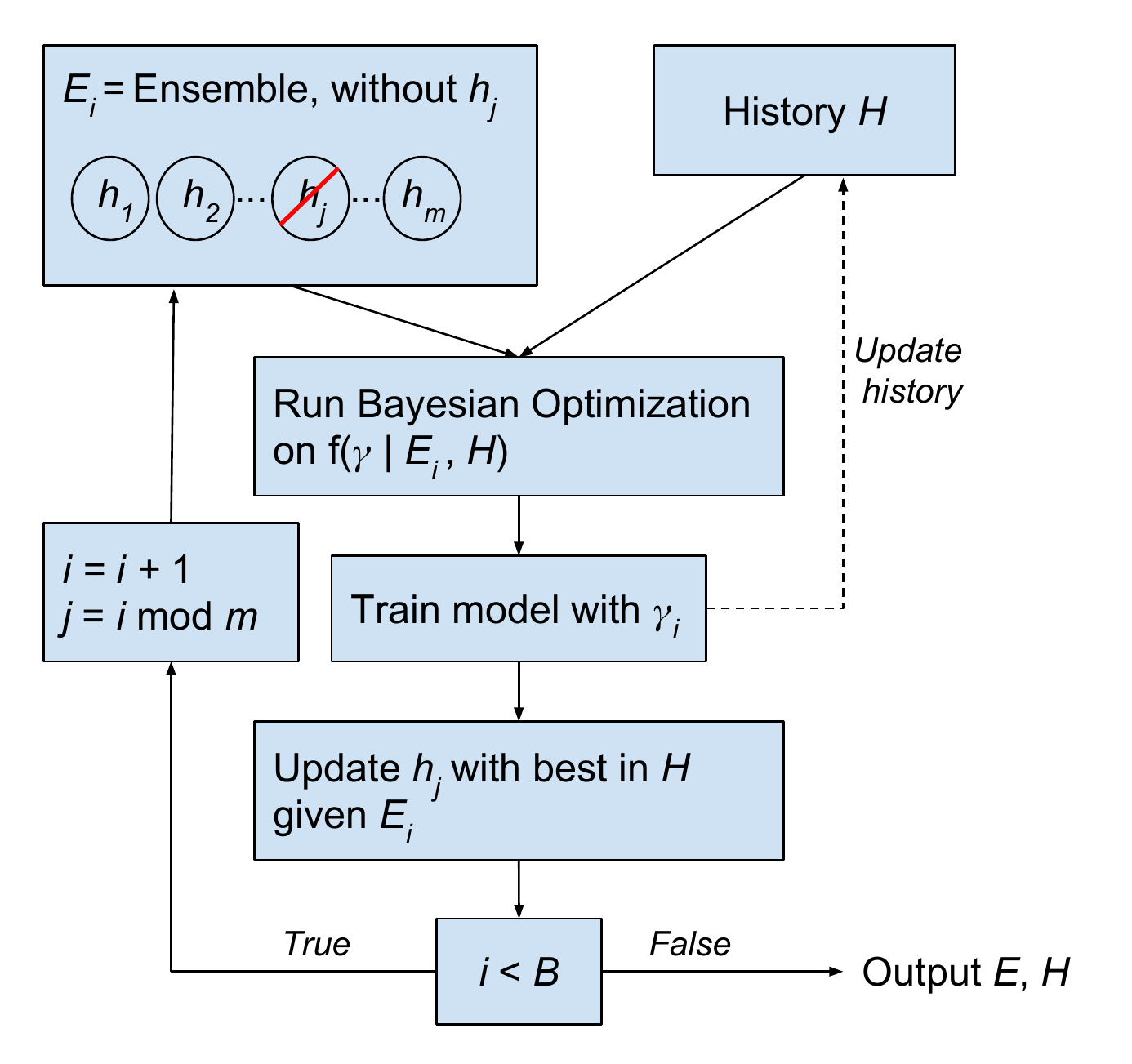}
    \caption{Schema of the Ensemble Optimization procedure.}
    \label{fig:eo_method}
\end{figure}

The problem defined above is straightforward as long as the ensemble is given beforehand. In this section we will tackle the problem of building and updating the ensemble as the optimization progresses. To make things simpler, an ensemble size $m$ will be fixed beforehand -- this number can be fine-tuned at the end of the optimization. Each iteration of the optimization procedure will contain two steps: first the evaluation as described in Section~\ref{sec:eo}, maintaining a fixed ensemble, and then an ensemble update step. Since members of the ensemble should be changed as the optimization moves forward and better models are found, a round-robin strategy will be used for the ensemble construction. The ensemble $E$ will in fact consist of $m$ fixed positions, and at every iteration $i$, the classifier at position $j = (i \mod m)$ will be removed from the ensemble before finding hyperparameters which minimize Equation~\ref{eq:ens_opt} -- effectively optimizing the classifier at this position for the given iteration. At the end of an iteration the ensemble is updated again greedily, selecting the new best classifier (it could be the same classifier or a better one). The whole procedure is described in Algorithm~\ref{alg:eo} and in Figure~\ref{fig:eo_method}.

\renewcommand{\algorithmicrequire}{\textbf{Input:}}
\renewcommand{\algorithmicensure}{\textbf{Output:}}
\algrenewcomment[1]{\hspace{0.05in} // #1}
\begin{algorithm}[t]
\caption{Ensemble Optimization Procedure}
\label{alg:eo}
\begin{algorithmic}[1]
\Require $\mathcal{X}_T, \mathcal{X}_V, B, m, A, \Gamma, L$
\Ensure $H$, history of models; $E$, the final ensemble
\State $H, G, E \gets \varnothing$
\For{$i \in {1,\ldots, B}$}
    \State $j \gets i \mod m$
    \State $E \gets E \setminus \{ h_j \}$
    \State $\mathbf{L}_i \gets \{L(E \cup h \vert \mathcal{X}_V)\}_{h \in H}$
    \State $f(\gamma \vert E) \gets \mathrm{BO}(G, \mathbf{L}_i)$ \Comment Fit model
    \State $\gamma_i \gets \argmax_{\gamma \in \Gamma} a(\gamma \vert f(\gamma \vert E))$ \Comment Next hypers
    \State $h_i \gets \mathrm{A}(\mathcal{X}_{T}, \gamma_i)$ \Comment Train model
    \State $G \gets G \cup \{\gamma_i\}$
    \State $H \gets H \cup \{h_i\}$
    \State $h_j \gets \argmin_{h \in H} L(E \cup \{h\})$\Comment New model at $j$
    \State $E \gets E \cup \{ h_j \} $ \Comment Update ensemble
\EndFor
\end{algorithmic}
\end{algorithm}

\begin{figure}[tb]
    \centering
    \includegraphics[width=0.48\textwidth]{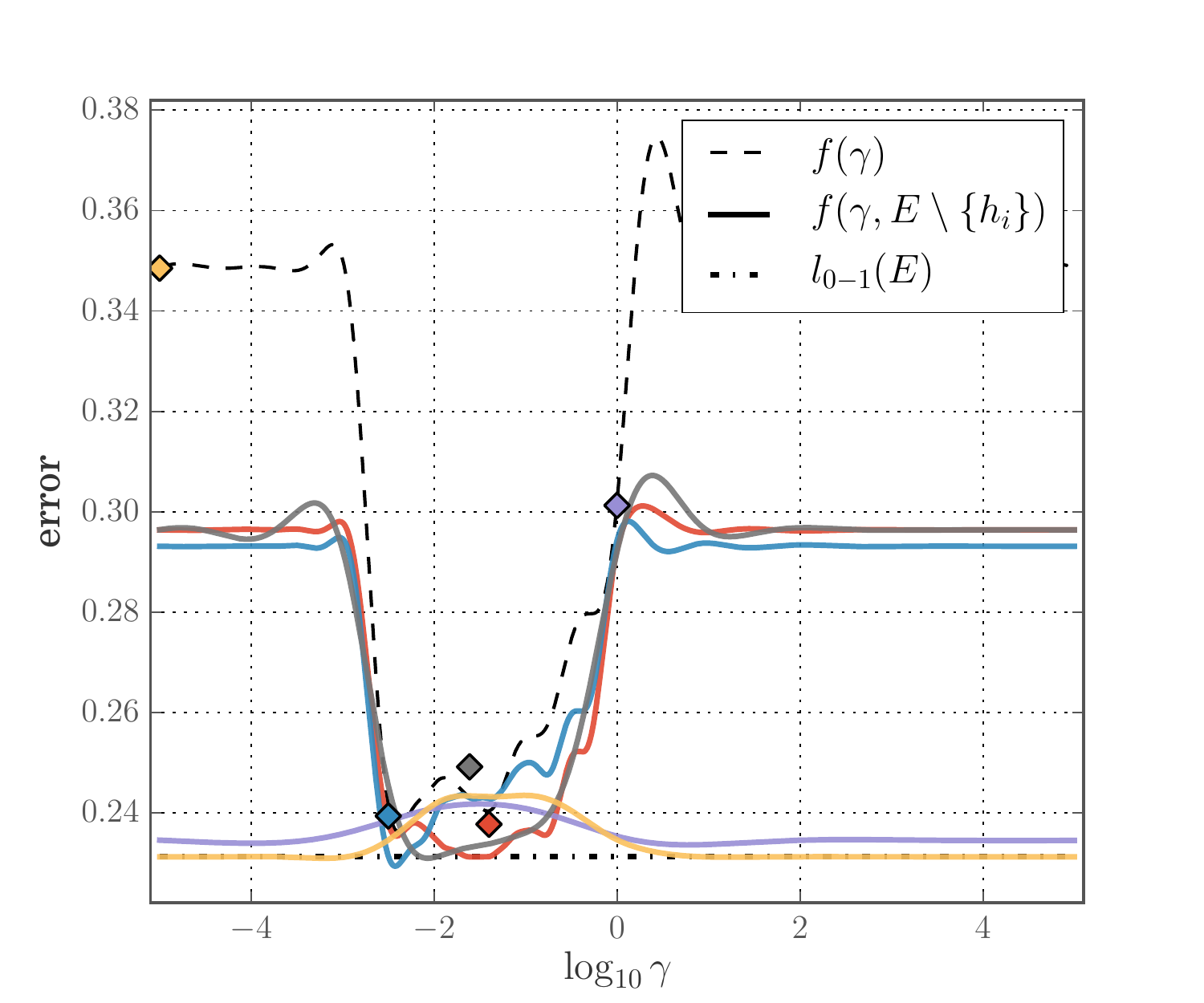}
    \caption{Example of objective function $f(\gamma \vert E \setminus \{h_{i}\})$ given a pool of 50 trained classifiers on a 1-D hyperparameter optimization problem. Each color represents the performance of an ensemble when removing its member $h_{i}$ represented by a diamond of the same color in the plot.}
    \label{fig:specialize}
\end{figure}

In addition, it is expected that some classifiers will specialize given the fixed state of a large part of the ensemble for each iteration. For instance, when replacing an individually strong classifier, another strong classifier will most likely be required. Figure~\ref{fig:specialize} shows an example of optimization on a one-dimensional hyperparameter space run for 50 iterations, where an ensemble of five classifiers was optimized. The ensemble is represented by the five diamonds and its generalization error is shown by the dotted and dashed line at the bottom of the figure. Then, each of the five members $i$ is independently removed and a Gaussian Process model is fit on the performance of an ensemble given the remaining models in the pool -- this corresponds to the five colored lines of Figure~\ref{fig:specialize}. We can see from this figure that the hyperparameters which minimize the ensemble error are different for each \emph{slot} in the ensemble, illustrating our concept.

\subsection{COMPUTATIONAL COMPLEXITY}

The \emph{computational overhead} of the proposed method comes mainly from the evaluation of the empirical error of ensembles. It is very small with regards to the cost of running most learning algorithms (which is usually quadratic or worse in the number of samples), and also with the cost of conditioning the probabilistic model on the observations (which is cubic in the number of iterations). The computation of the empirical error of ensembles takes place in step 5 in Algorithm~\ref{alg:eo}. Given an ensemble of size $m$, a validation dataset of size $n$, and a history of trained classifiers of size $t$, the complexity of this step is $O(t(mn + n)) = O(tmn)$ since it requires one pass over all models in the history, and for each of those the combination of the classifiers through majority voting ($mn$) and the computation of the empirical error $n$.

\subsection{LOSS FUNCTION}

The objective function defined in Equation~\ref{eq:ens_opt} contains a loss function, which up until now referred to the empirical loss of the ensemble, or the zero-one loss.
However, the zero-one loss contains a strong discontinuity and can result in optimization procedures failing due to the majority voting combination. For instance, if all classifiers of the ensemble are wrong on some instances, replacing one of those poor classifiers with a better one will not make the ensemble correctly classify those instances, resulting in the same performance with regards to the objective function, even though this classifier would be a good choice for the ensemble.

The performance of ensembles will be considered with regards to their classification \emph{margin}, which will let us derive a more suitable loss function~\citep{Schapire2012}. Given an ensemble of classifiers $E$ outputting label predictions on a binary problem $\mathcal{Y} \in \{-1, 1\}$, the \emph{normalized} margin for a sample $\{x, y\}$ is defined as follows:
\begin{equation}
M(E, x, y) = \frac{1}{\vert E \vert} \sum_{h \in E} y h(x).
\end{equation}
The normalized margin $M \in [-1, 1]$ takes the value $1$ when all classifiers of the ensemble correctly classify the sample $x$, $-1$ when all the classifiers are wrong, and somewhere in between otherwise. In the case of multi-class problems, predictions of classifiers can be brought back to a binary domain by attributing $1$ for a correct classification and $-1$ for a misclassification. The margin becomes:
\begin{equation}
M_{mc}(E, x, y) = \frac{1}{\vert E \vert} \sum_{h \in E} [1 - 2 l_{0-1}\left(h(x), y\right)].
\end{equation}

We will now derive some loss functions from the margin. The margin itself could be the objective, since it is desirable that the margin of the ensemble be high. It must be rescaled to really become a loss function, giving a margin-based loss function:
\begin{equation}
\label{eq:margin_loss}
     l_M(E, x, y) = \frac{1 - M(E, x, y)}{2}.
 \end{equation}
This loss function should not be used to optimize an ensemble because it is directly maximized only by the accuracy of individual classifiers. In other words, given a validation dataset $\mathcal{X}_V$ and a set of classifiers $H$ to evaluate, the classifier minimizing Equation~\ref{eq:margin_loss} is always the classifier with the lowest empirical error on its own, without regards to the ensemble performance. Therefore, the loss function must not give the same weight to all classifiers without regards to ensemble performance, while also being smooth. A loss function which achieves this is the margin-based loss function taken to the power of two:
\begin{equation}
l_{M^2}(E, x, y) =  \frac{(1 - M(E, x, y))^2}{4}.
\end{equation}
This places a higher emphasis on samples misclassified by the ensemble, and decreases the importance of samples as the margin grows closer to 1. Since it meets the required properties, the squared margin loss function will be used as the ensemble loss function in this work.


\begin{figure}[tb]
    \centering
    \includegraphics[width=0.48\textwidth]{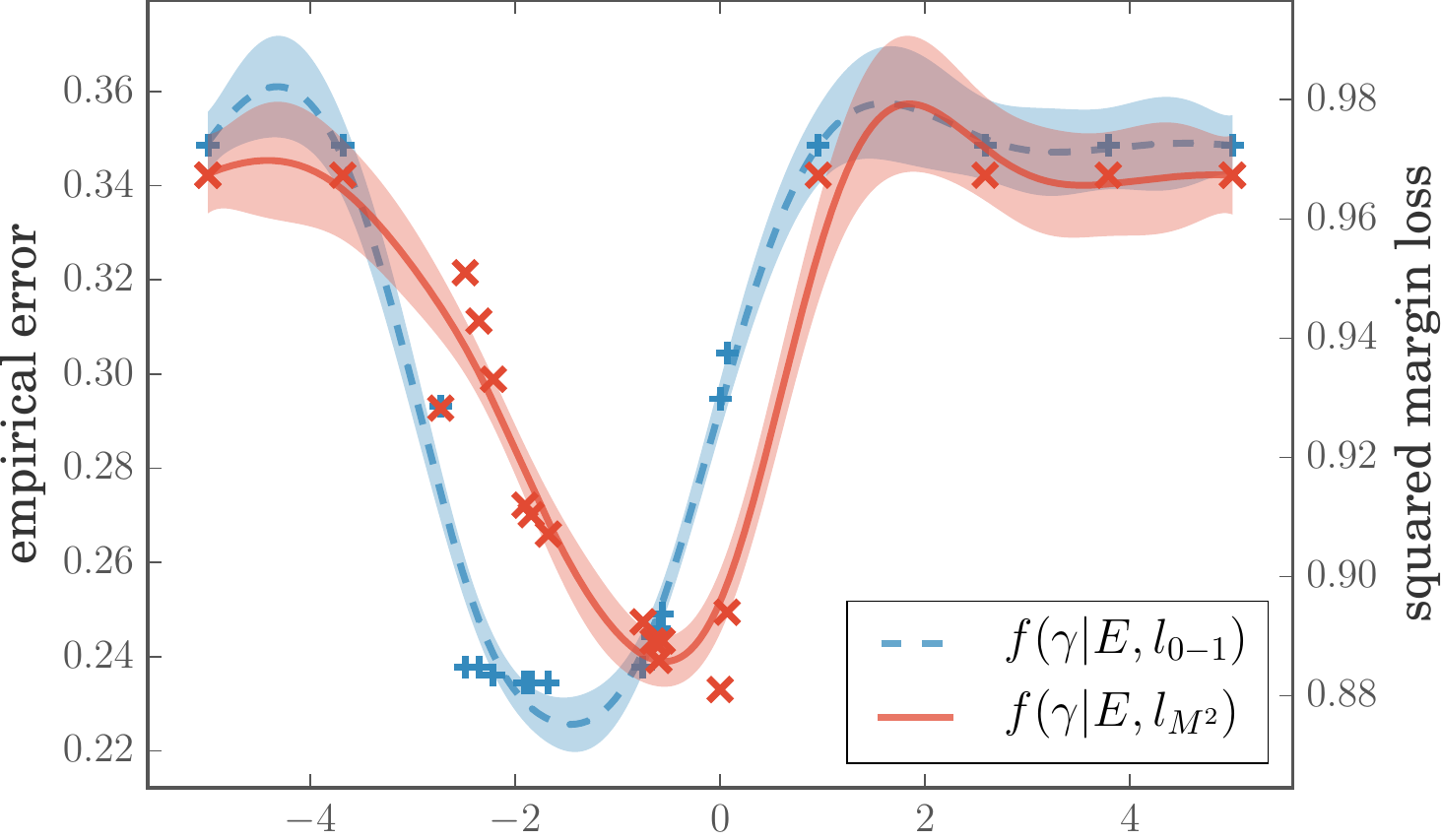}
    \caption{Examples of ensemble loss functions for a 1-D hyperparameter optimization. Blue curve legend is on the left, red curve legend is on the right.}
    \label{fig:loss_example}
\end{figure}

Figure~\ref{fig:loss_example} shows an example of the two loss functions discussed in this section, the zero-one loss and the squared margin loss, applied on the same ensemble. Both losses have different scales, with the empirical error scale on the left and the squared margin scale on the right of the figure. We can see that these two loss functions lead to different optimal hyperparameters given the same ensemble. In other words, the hyperparameters minimizing the objective function according to the models of $f(\gamma | E)$ are different with the two loss functions.

Considering that a loss function other than the empirical error will be used during the optimization, it will probably be beneficial to rebuild an ensemble from scratch using the final pool of classifiers trained once the optimization is over. Hence, a post-hoc ensemble generation will be performed after the hyperparameter optimization.

\section{EXPERIMENTS}
\label{sec:exp}

\begin{table}[tb]
    \caption{Benchmarking datasets and abbreviations}
    \label{tab:datasets}
    \centering
\begin{small}
\begin{tabular}{lrrr}
    \toprule
    Dataset & Instances & Features & Classes \\
    \midrule
    Adult (adlt) & 48,842 & 14 & 2 \\
    Bank (bnk) & 4,521 & 16 & 2\\
    Car (car) & 1,728 & 6 & 4\\
    Chess-krvk (ches) & 28,506 & 6 & 18 \\
    Letter (ltr) & 20,000 & 16 & 26 \\
    Magic (mgic) & 19,020 & 10 & 2 \\
    Musk-2 (msk) & 6,598 & 166 & 2 \\
    Page-blocks (p-blk) & 5,473 & 10 & 5 \\
    Pima (pim) & 768 & 8 & 2 \\
    Semeion (sem) & 1,593 & 256 & 10 \\
    Spambase (spam) & 4,601 & 57 & 2 \\
    Stat-german-credit (s-gc) & 1,000 & 24 & 2\\
    Stat-image (s-im) & 2,310 & 18 & 7 \\
    Stat-shuttle (s-sh) & 58,000 & 9 & 7 \\
    Steel-plates (s-pl) & 1,941 & 27 & 7 \\
    Titanic (tita) & 2,201 & 3 & 2 \\
    Thyroid (thy) & 7,200 & 21 & 6 \\
    Wine-quality-red (wine) & 1,599 & 11 & 6 \\ \bottomrule
\end{tabular}
\end{small}
\end{table}

We showcase the performance of our ensemble optimization approach on three different problems. The first two problems consist of different algorithms and hyperparameter spaces evaluated on the same benchmark of medium datasets available from the UCI repository, presented in Table~\ref{tab:datasets}. For every repetition, a different hold-out testing partition was sampled with $33\%$ of the total dataset size (unless the dataset had a pre-specified testing split, in which case it was used for all repetitions). The remaining instances of the dataset were used to complete a 5-fold cross-validation procedure. Final models are retrained on all the data available for training and validation.

In every case, the prior on the objective function is a Gaussian Process (GP) with Mat\'{e}rn-52 kernel using automatic relevance determination\ftnote{Code from \url{http://github.com/JasperSnoek/spearmint}.}. The noise, amplitude, and length-scale parameters are obtained through slice sampling~\citep{Snoek2012}. The slice sampling of GP hyperparameters for ensemble optimization must be reinitialized at every iteration given that different ensembles $E$ can change the properties of the optimized function drastically. The acquisition function is the Expected Improvement over the best solution found so far.
The compared methods and their abbreviated names are the following:
\begin{itemize}
    \item Classical Bayesian optimization (BO-best). It returns a single model selected with argmin on validation performance~\citep{Snoek2012}.
    \item Post-hoc ensemble constructed from the pool of classifiers with Bayesian optimization (BO-post). The post-hoc ensemble is initiated by picking the three best classifiers from the pool before proceeding with forward greedy selection -- this form of warm starting is recommended in~\citep{Caruana2004} to reduce overfitting.
    \item The proposed ensemble optimization method using the squared margin loss function (EO).
    \item Post-hoc ensemble constructed from the pool of classifiers generated by ensemble optimization (EO-post). The same post-hoc procedure as with BO-post is executed.
\end{itemize}


The hyperparameter spaces can contain continuous, discrete, and categorical parameters (e.g., base classifier or kernel choice). In the case of categorical and discrete parameters, they are represented using a continuous parameter which is later discretized. This does not deal with the fact that hyperparameter spaces of different classifiers are disjoint and should not be modeled jointly, but since all the compared methods are using this same technique, the comparison is fair. 

\subsection{SVM SEARCH SPACE}
\label{sec:svm_exp}

The models used in this benchmark are SVM models, and the parameterization includes the choice of the kernel along with the various hyperparameters needed per kernel. The hyperparameter space $\Gamma$ optimized can be described as follows:
\begin{itemize}
     \item One hyperparameter for the kernel choice: linear, RBF, polynomial, or sigmoid;
     \item Configurable error cost $C \in [10^{-5}, 10^5]$ (for all kernels);
     \item RBF and sigmoid kernels both have a kernel width parameter $\gamma_{RBF} \in [10^{-5}, 10^5]$;
     \item Polynomial kernel has a degree parameter $d \in [1, 10]$;
     \item Sigmoid and polynomial kernels both have an intercept parameter, $c \in [10^{-2}, 10^2]$.
 \end{itemize}

All compared approaches optimized the same search space. Each method is given a budget of $B=200$ iterations, or 200 hyperparameter tuples tested, to optimize hyperparameters with a 5-fold cross-validation procedure. The ensemble selection stage exploits this cross-validation procedure, considering the next classifier which reduces the most the generalization error over all the cross-validation folds. Selected hyperparameters are retrained on the whole training and validation data, and combined directly on the testing split to generate the generalization error values presented in this section. The ensemble optimization method is run with an ensemble size $m=12$. This ensemble size was selected empirically and may not be optimal. Future work could investigate strategies to dynamically size the ensemble as the optimization progresses, with no fixed limit.

The generalization error on the test split for the selected methods is presented in Table~\ref{tab:medium_err_5fold}, averaged over 10 repetitions. The last column shows the ranks of each method averaged over all datasets, where the best rank is 1 and the worst rank is 4.

\begin{table*}[tb]
\centering
\caption{Generalization error on SVM hyperparameter space, averaged over 10 repetitions, 5-fold cross-validation. Last column shows the rank of methods averaged over all datasets.}
\label{tab:medium_err_5fold}
\setlength{\tabcolsep}{3pt}
\resizebox{\textwidth}{!}{
\begin{tabular}{lcccccccccccccccccc|c}
\toprule
         &   adlt &   bnk &   car &   ches &   ltr &   mgic &   msk &   p-blk &   pim &   sem &   spam &   s-gc &   s-im &   s-sh &   s-pl &   thy &   tita &   wine &   Ranks \\
\midrule
 BO-best &  15.52 & 10.67 &  1.27 &  16.86 &  2.45 &  12.49 &  0.29 &    3.06 & 25.52 &  4.43 &   6.47 &  23.20 &   3.57 &   0.10 &  23.91 &  3.09 &  20.59 &  35.28 &    3.39 \\
 BO-post &  15.38 & 10.71 &  1.56 &  16.72 &  2.50 &  12.21 &  0.28 &    3.01 & 25.65 &  4.37 &   6.47 &  23.45 &   2.94 &   0.08 &  22.58 &  3.17 &  20.59 &  35.09 &    2.81 \\
 EO      &  15.39 & 10.44 &  0.81 &  15.06 &  2.34 &  12.18 &  0.30 &    3.14 & 23.70 &  4.58 &   6.45 &  23.05 &   2.73 &   0.09 &  22.61 &  2.51 &  20.27 &  33.29 &    1.89 \\
 EO-post &  15.27 & 10.60 &  0.95 &  15.08 &  2.36 &  12.21 &  0.28 &    2.97 & 24.03 &  4.40 &   6.36 &  23.40 &   2.55 &   0.09 &  22.63 &  2.69 &  20.57 &  33.70 &    1.92 \\
\bottomrule
\end{tabular}
}
\end{table*}

\begin{table}[tb]
\centering
\caption{Wilcoxon pairwise test $p$-values for the SVM hyperparameter space. Bold entries highlight significant differences ($p \leq 0.05$) and parentheses are added when method at row $i$ is worse than the method at column $j$ according to ranks.}
\label{tab:svm_wilcoxon_pw_5f}
\setlength{\tabcolsep}{3pt}
\begin{tabular}{lcccc}
\toprule
&1&2&3&4 \\ \midrule
1 - BO-best & -- & (0.33) & (\textbf{0.00}) & (\textbf{0.00})\\
2 - BO-post & 0.33 & -- & (\textbf{0.01}) & (\textbf{0.00})\\
3 - EO & \textbf{0.00} & \textbf{0.01} & -- & 0.21\\
4 - EO-post & \textbf{0.00} & \textbf{0.00} & (0.21) & -- \\ \bottomrule
\end{tabular}
\end{table}

A Wilcoxon signed-rank test is used to measure the statistical significance of the results. The Wilcoxon signed-rank test is a strong statistical test for comparing methods across multiple datasets, which is an nonparametric version of the Student's $t$-test that does not assume normal distributions and is less sensitive to outliers~\citep{Demsar2006}.
The input for the Wilcoxon test is the generalization error of a method $i$ on each dataset $d$, averaged across the $R$ repetitions:
\begin{equation}
\mathbf{e}_i = \{ \frac{1}{R} \sum_{r=1}^R e_{i,d,r} \}_d,
\end{equation}
where $e_{i,d,r}$ is the generalization error produced by method $i$ on dataset $d$ at repetition $r$. The Wilcoxon test is then computed for all pairs of methods $(\mathbf{e}_i, \mathbf{e}_j)$. The results of this procedure are shown in Table~\ref{tab:svm_wilcoxon_pw_5f}.


From Table~\ref{tab:medium_err_5fold} we can see that it is beneficial to build an ensemble from the output of a classical hyperparameter optimization, as seen by the lower rank of BO-post with regards to BO-best. However, the performance improvement is not shown to be significant according to the Wilcoxon test. Both the ensemble optimization methods seem to outperform classical Bayesian optimization strategies in terms of rankings. The Wilcoxon test shows that EO and EO-post both performed significantly better than BO-best and BO-post. It should be noted that there is no significant difference between EO and EO-post, highlighting that there was not a significant gain from the post-hoc ensemble construction. \cite{Caruana2004} presented some strategies to reduce overfitting in the forward greedy procedure -- such as bagging from the pool of models -- which could be considered in order to achieve more with the same pool, although this is left for future work.

\begin{figure}[tb]
    \centering
    \includegraphics[width=0.4\textwidth]{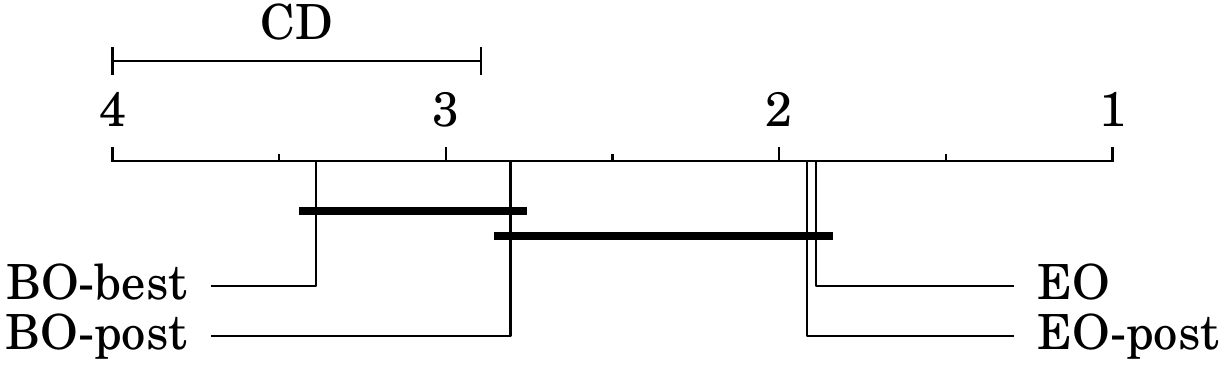}
    \caption{Methods by rank and significant differences according to a post-hoc Nemenyi test with significance level at $p=0.05$ for the SVM hyperparameter space.}
    \label{fig:nemenyi_svm}
\end{figure}

Another test which can be used to assess the performance of the evaluated methods is the Friedman test with post-hoc tests on classifier ranks averaged across datasets. A Friedman test with the four methods presented in this section finds a significant difference between them with a $p$-value of $\num{5.5e-4}$. The Friedman test is then usually followed by a post-hoc test to measure whether the difference in ranks is above a critical difference level, such as the Nemenyi test~\citep{Demsar2006}. Figure~\ref{fig:nemenyi_svm} shows the results of such a test, with methods linked by bold lines being found not significantly different by the test for a significance level of $p=0.05$. The Nemenyi post-hoc test gives a more visual insight as to what is going on, but it is more sensitive to the pool of tested methods -- the outcome of the test can change if new methods are inserted in the experiments. According to this test, EO and EO-post are both significantly different from BO-best, meaning that ensemble optimization is significantly better than the single best classifier returned by Bayesian optimization.

\subsection{SCIKIT-LEARN MODELS SEARCH SPACE}

The same methods as in the previous section were repeated for a different search space consisting of multiple base algorithms, all available from \texttt{scikit-learn}\ftnote{Available at \url{http://scikit-learn.org/}.}. The models and their hyperparameters are \emph{K nearest neighbors} with the number of neighbors \texttt{n\_neighbors} in $[1, 30]$; \emph{RBF SVM} with the penalty \texttt{C} logarithmically in $[10^{-5}, 10^5]$ and the width of the RBF kernel \texttt{$\gamma_{RBF}$} logarithmically in $[10^{-5}, 10^5]$; \emph{linear SVM} with the penalty \texttt{C} logarithmically scaled in $[10^{-5}, 10^5]$; \emph{decision tree} with the maximal depth \texttt{max\_depth} in $[1, 10]$, the minimum number of examples in a node to split \texttt{min\_samples\_split} in $[2, 100]$, and the minimum number of training examples in a leaf \texttt{min\_samples\_leaf} in $[2, 100]$; \emph{random forest} with the number of trees \texttt{n\_estimators} in $[1, 30]$, the maximal depth \texttt{max\_depth} in $[1, 10]$, the minimum number of examples in a node to split \texttt{min\_samples\_split} in $[2, 100]$, and the minimum number of training examples in a leaf \texttt{min\_samples\_leaf} in $[2, 100]$; \emph{AdaBoost} with the number of weak learners \texttt{n\_estimators} in $[1, 30]$; \emph{Gaussian Naive Bayes (GNB)} and \emph{Linear Discriminant Analysis (LDA)} both without any hyperparameters; and \emph{Quadratic Discriminant Analysis (QDA)} with the regularization \texttt{reg\_param} logarithmically in $[10^{-3}, 10^3]$.

The set of hyperparameter optimization strategies is the same as in the previous section, and the maximum number of iterations $B$ is set to 100. Tables~\ref{tab:sklearn_err_5fold} and~\ref{tab:sklearn_wilcoxon_pw_5f} show the same metrics and Wilcoxon pairwise tests as introduced in Section~\ref{sec:svm_exp}.

\begin{table*}
\centering
\caption{Generalization error on the scikit-learn hyperparameter space, averaged over 10 repetitions, 5-fold cross-validation. Last column shows the rank of methods averaged over all datasets.}
\label{tab:sklearn_err_5fold}
\setlength{\tabcolsep}{3pt}
\resizebox{\textwidth}{!}{
\begin{tabular}{lcccccccccccccccccc|c}
\toprule
         &   adlt &   bnk &   car &   ches &   ltr &   mgic &   msk &   p-blk &   pim &   sem &   spam &   s-gc &   s-im &   s-sh &   s-pl &   thy &   tita &   wine &   Ranks \\
\midrule
 BO-best &  14.70 & 11.04 &  4.77 &  25.73 &  4.72 &  12.06 &  2.13 &    3.41 & 23.25 &  8.82 &   5.80 &  23.15 &   3.72 &   0.12 &  26.91 &  1.19 &  22.74 &  35.96 &    3.36 \\
 BO-post &  14.62 & 10.53 &  4.80 &  25.73 &  4.72 &  11.99 &  2.18 &    3.26 & 23.38 &  8.82 &   5.53 &  23.15 &   3.64 &   0.11 &  26.16 &  1.20 &  22.74 &  35.71 &    3.11 \\
 EO      &  14.35 & 10.30 &  0.81 &  19.39 &  2.81 &  12.38 &  0.28 &    2.80 & 24.09 &  4.37 &   5.02 &  22.75 &   2.84 &   0.08 &  22.69 &  1.21 &  20.66 &  35.19 &    1.67 \\
 EO-post &  14.32 & 10.39 &  1.01 &  20.47 &  2.76 &  12.48 &  0.28 &    2.94 & 23.38 &  4.37 &   5.24 &  22.90 &   2.99 &   0.07 &  23.50 &  1.10 &  21.20 &  35.47 &    1.86 \\
\bottomrule
\end{tabular}
}
\end{table*}

\begin{table}
\centering
\caption{Wilcoxon pairwise tests $p$-values for the scikit-learn hyperparameter space. Bold entries highlight significant differences ($p \leq 0.05$) and parentheses are added when method at row $i$ is worse than the method at column $j$ according to ranks.}
\label{tab:sklearn_wilcoxon_pw_5f}
\setlength{\tabcolsep}{3pt}
\begin{tabular}{lcccc}
\toprule
&1&2&3&4 \\ \midrule
1 - BO-best & -- & (\textbf{0.05}) & (\textbf{0.00}) & (\textbf{0.00})\\
2 - BO-post & \textbf{0.05} & -- & (\textbf{0.00}) & (\textbf{0.00})\\
3 - EO & \textbf{0.00} & \textbf{0.00} & -- & \textbf{0.03}\\
4 - EO-post & \textbf{0.00} & \textbf{0.00} & (\textbf{0.03}) & -- \\ \bottomrule
\end{tabular}
\end{table}

Conclusions are similar for this hyperparameter space, whereas the generation of a post-hoc ensemble is again shown to be beneficial to the generalization accuracy for both BO and EO. In this case the post-hoc ensemble for BO is found significantly better than the single best classifier according to the Wilcoxon test. The overall best method is EO, which significantly outperforms all other methods, including EO-post, as seen in Table~\ref{tab:sklearn_wilcoxon_pw_5f}.  For some datasets, the ensemble optimization procedure achieves a large drop in the generalization error, see for example datasets letter (ltr), musk-2 (msk) and semeion (sem) in Table~\ref{tab:sklearn_err_5fold}.

\begin{figure}[tb]
    \centering
    \includegraphics[width=0.4\textwidth]{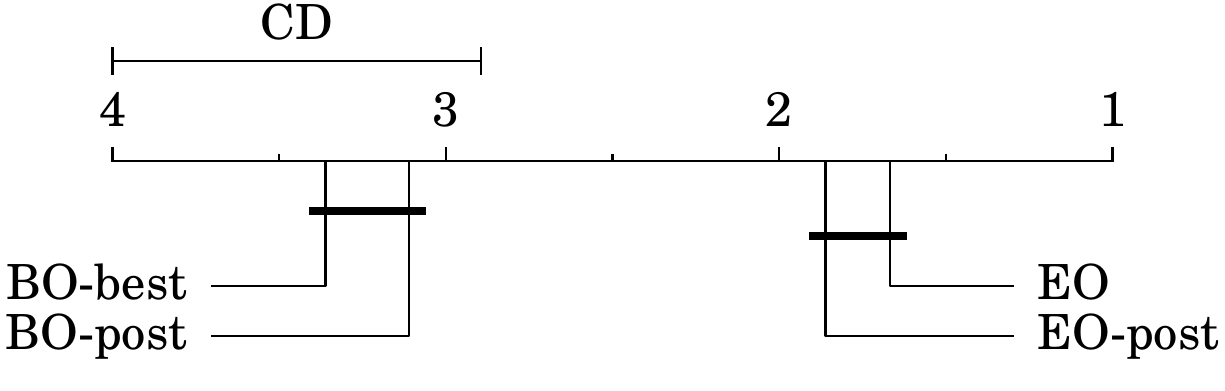}
    \caption{Methods by rank and significant differences according to a post-hoc Nemenyi test with significance level at $p=0.05$ for the scikit-learn hyperparameter space.}
    \label{fig:nemenyi_sklearn}
\end{figure}

A Friedman test on this suite of experiments is again run and found significant with a $p$-value of $\num{1.5e-5}$. Figure~\ref{fig:nemenyi_sklearn} shows the results of the Nemenyi post-hoc test, with methods linked by bold lines being found not significantly different by the test for a significance level of $p=0.05$. According to this test, EO-post and EO are significantly different from both BO and BO-best, meaning that ensemble optimization approaches significantly outperform both Bayesian optimization baselines.

\subsection{CONVOLUTIONAL NEURAL NETWORKS}

Lastly, we evaluated the performance of our approach when fine-tuning the parameters of a convolutional neural network for the CIFAR-10 dataset. In order to have a reproducible baseline, the \texttt{cuda-convet} implementation was used with the reference model files given which achieves 18\% generalization error on the testing dataset\footnote{Code available at \url{https://code.google.com/archive/p/cuda-convnet/} and network configuration file used is \texttt{layers-18pct.cfg}.}.
One batch of the training data was set aside for validation (batches 1-4 used for training, 5 for validation, and 6 for testing). Performance of the baseline configuration on the given training batches was around $22.4\% \pm 0.9$ for 250 epochs of training. The parameters optimized were the same as in~\citep{Snoek2012}, namely the learning rates and weight decays for the convolution and softmax layers, and the parameters of the local response normalization layer (size, power and scale). The number of training epochs was kept fixed at 250.

\begin{figure}[tb]
    \centering
    \includegraphics[width=0.45\textwidth]{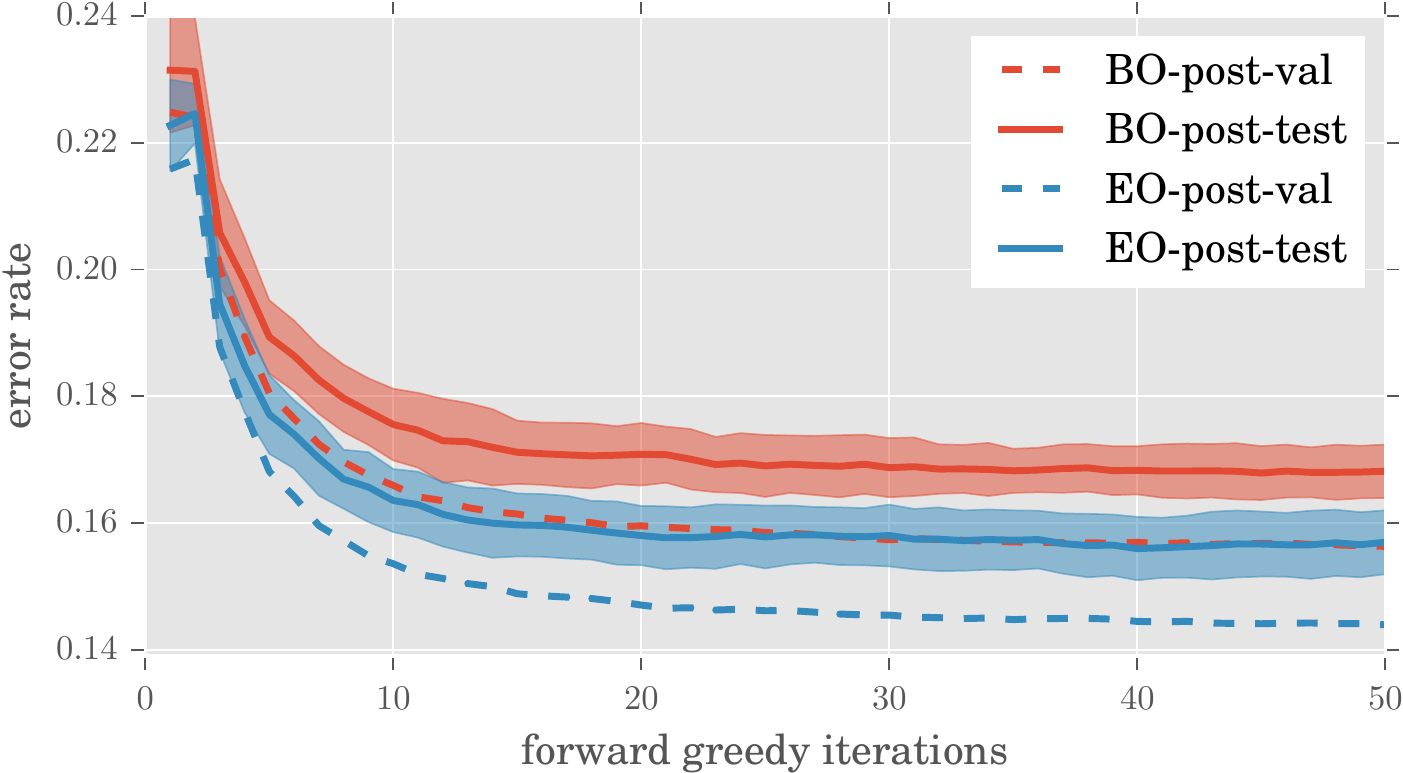}
    \caption{Generalization errors of a post-hoc ensemble with classical Bayesian optimization (BO-post) and a post-hoc ensemble generated from our ensemble optimization approach (EO-post) on the CIFAR-10 dataset with regards to the number of classifiers in the final ensemble. Results averaged over 10 repetitions.}
    \label{fig:cnn_fwgreedy}
\end{figure}

We computed 10 repetitions of a standard Bayesian optimization and our proposed ensemble optimization with ensemble size $m=7$, both with a budget of $B=100$ hyperparameter tuples to evaluate. Figure~\ref{fig:cnn_fwgreedy} shows the performance of ensembles generated from both pools of classifiers with a post-hoc ensemble generation. In order to limit overfitting, the first three models of each ensemble were selected directly based on accuracy, as suggested in~\citep{Caruana2004}. In both cases, the ensemble size benefits the generalization accuracy, although the classifiers generated by the ensemble optimization procedure do perform slightly better. The difference in generalization error between BO-post and EO-post at the last iteration is found significant by a Wilcoxon test with a $p$-value of 0.005.
Further work should investigate strategies to use the remaining validation data once the models are chosen, to further improve generalization accuracy.

\section{CONCLUSION}

In this work, we presented a methodology to achieve Bayesian optimization of ensembles through hyperparameter tuning. We tackle the various challenges posed by ensemble optimization in this context, and the result is an optimization strategy that is able to exploit trained models and generate better ensembles of classifiers at the computational cost of a regular hyperparameter optimization.

We showcase the performance of our approach on three different problem suites, and in all cases show a significant difference in generalization accuracy between our approach and post-hoc ensembles built on top of a classical hyperparameter optimization, according to Wilcoxon signed-rank tests. This is a strong validation of our method, especially considering that it involves little extra computation.


\subsubsection*{Acknowledgements}

This research benefitted from the computing resources provided by Calcul Qu\'{e}bec, Compute Canada and Nvidia. We would also like to thank Annette Schwerdtfeger for proofreading this paper.

\renewcommand{\refname}{}

{\def\section*#1{}
\subsubsection*{References}
\printbibliography
}

\end{document}